\documentclass{article}

\usepackage[final]{neurips_2020}

\usepackage[utf8]{inputenc} 
\usepackage[T1]{fontenc}    
\usepackage{hyperref}       
\usepackage{url}            
\usepackage{booktabs}       
\usepackage{amsfonts}       
\usepackage{nicefrac}       
\usepackage{microtype}      
\usepackage{graphicx}
\usepackage{amsmath}

\title{Multi-stage, multi-swarm PSO for joint optimization of well placement and control}

\author{%
  Ajitabh Kumar\\
  Visage Technology\\
}

\begin{document}

\maketitle

\begin{abstract}
Evolutionary optimization algorithms, including particle swarm optimization (PSO), have been successfully applied in oil industry for production planning and control. Such optimization studies are quite challenging due to large number of decision variables, production scenarios, and subsurface uncertainties. In this work, a multi-stage, multi-swarm PSO (MS2PSO) is proposed to fix certain issues with canonical PSO algorithm such as premature convergence, excessive influence of global best solution, and oscillation.

Multiple experiments are conducted using Olympus benchmark to compare the efficacy of algorithms. Canonical PSO hyperparameters are first tuned to prioritize exploration in early phase and exploitation in late phase. Next, a two-stage multi-swarm PSO (2SPSO) is used where multiple-swarms of the first stage collapse into a single swarm in the second stage. Finally, MS2PSO with multiple stages and multiple swarms is used in which swarms recursively collapse after each stage. Multiple swarm strategy ensures that diversity is retained within the population and multiple modes are explored. Staging ensures that local optima found during initial stage does not lead to premature convergence. Optimization test case comprises of 90 control variables and a twenty year period of flow simulation. It is observed that different algorithm designs have their own benefits and drawbacks. Multiple swarms and stages help algorithm to move away from local optima, but at the same time they may also necessitate larger number of iterations for convergence. Both 2SPSO and MS2PSO are found to be helpful for problems with high dimensions and multiple modes where greater degree of exploration is desired.
\end{abstract}

\section{Introduction}
Oilfield planning, development and production involves comparing different scenarios, and then selecting optimal design. These decisions have direct impact on the commercial viability and success of any project. Such studies are done using reservoir simulation workflow in which different scenarios are modeled to estimate return on investment.

Evolutionary optimization algorithms, including particle swarm optimization (PSO), have been extensively used in such studies as they are population based and do not require gradient information [4, 5, 6]. PSO has successfully been used in many different applications including oilfield production optimization [1, 3, 7, 11]. Large scale oilfield production optimization has many design and control variables, while objective function may have multiple modes. High dimensionality and multi-modality makes divergent goals of exploration and convergence difficult to achieve. Canonical PSO has high dependence on the best particle of population, and hence suffers from premature convergence [19].

Different approaches have been explored in this work to promote exploration and mitigate premature convergence of PSO. This includes tuning hyperparameters of canonical PSO and using multiple swarms and stages. The main contributions of this work are:
\begin{itemize}
  \item Performance of canonical PSO using different hyperparameter settings are compared using Olympus benchmark.
  \item Two-stage PSO (2SPSO) is proposed where multiple-swarm of first stage reduces to a single swarm in second stage.
  \item Multi-stage, multi-swarm PSO (MS2PSO ) is proposed where swarms recursively collapse into a single swarm.
\end{itemize}

\section{Approach}

Olympus benchmark case is a synthetic simulation model based on a typical North Sea offshore oilfield, and is used in this work to compare the effectiveness of different optimization algorithms [2]. Field is a 9km by 3km by 50m in size with minor faults and limited aquifer support. Simulation model has 11 producer and 7 injector wells, all on bottom hole pressure constraint (Figure~\ref{olympus_model}). As a field redevelopment exercise, location of three production wells are searched using optimization workflow while others are treated as existing wells. Well location is defined using two 3-D coordinate sets indicating heel and toe for any well, thus having total six variables for a well. Bottom-hole pressure (BHP) value for production control is also optimized for all the 18 wells at four predetermined times during the 20 year simulation period. Thus, there are total 90 control variables which are optimized for the objective function under consideration. Weighted sum of cumulative fluid (WCF) is used as objective and is defined as:
\begin{figure}[t]
  \centering
  \includegraphics[width=\textwidth]{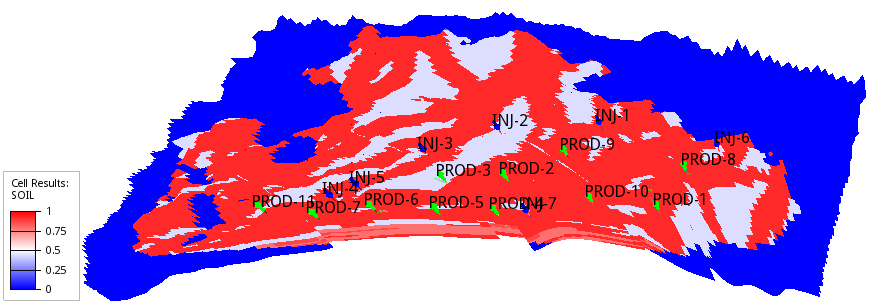}
  \caption{Olympus simulation model overview: initial oil saturation and well locations}
  \label{olympus_model}
\end{figure}
\begin{equation}
WCF = Q_{op} - 0.1 * (Q_{wp}+Q_{wi})
\end{equation}
where $Q_{op}$ is total oil produced, $Q_{wp}$ is total water produced, and $Q_{wi}$ is total water injected.

\section{Algorithm}
Swarm intelligence refers to global patterns emerging from simple interactions among population. Algorithmic rules at micro level lead to social interaction at meso level, which then further leads to collective behaviour at macro level. Thus, it is imperative to understand the impact of micro level rules on meso level particle interactions [20]. Self organization of swarm could be understood based on type of information, how particles use information, and flow of information within system (Figure~\ref{information}). PSO algorithm, and flocking in general, is sematectonic as position of particle itself drives learning, information travels by diffusion, and follow-through is used to continuously drive motion [23]. In comparison, self-organization in foraging is marker-based using pheromones emitted by the individuals, serendipitous in that an individual wanders randomly until it crosses by chance, and follow-through as distributed structures or trails to be continuously followed by individuals.

Canonical PSO algorithm derives swarm intelligence using particle position and velocity which are calculated as:
\begin{figure}[t]
  \centering
  \includegraphics[width=\textwidth]{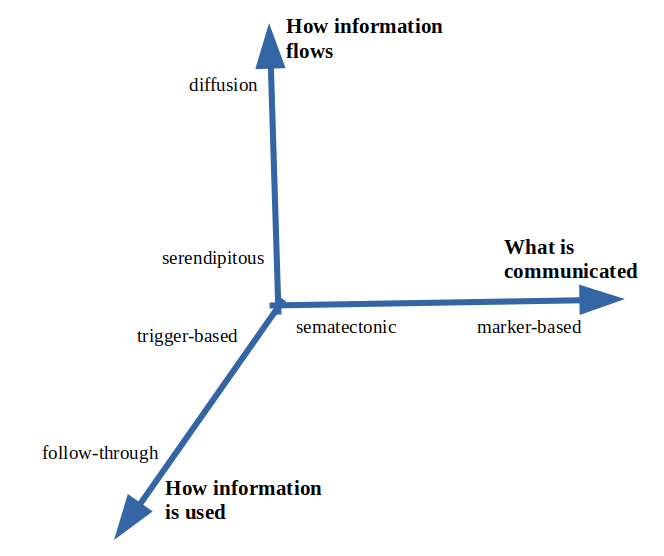}
  \caption{Information transfer and interaction mechanisms in self-organizing system (adapted from Mamei et al. 2006)}
  \label{information}
\end{figure}

\begin{equation}
v_i^d = \omega v_i^d + c_1r_1^d(pbest_i^d-x_i^d) + c_2r_2^d(gbest_i^d-x_i^d) \\
\end{equation}
\begin{equation}
x_i^d = x_i^d + v_i^d
\end{equation}
where $v_i^d$ and $x_i^d$ are velocity and position respectively of particle $i$ in dimension $d$, $\omega$ is the inertia weight, $c_1$ and $c_2$ are the cognitive and social acceleration coefficients respectively, $r_1^d$ and $r_2^d$ are independent uniformly distributed random numbers in [0,1], $pbest_i^d$ is the personal best for particle $i$, and $gbest_i^d$ is the global best position. PSO algorithm in this form tends to have premature convergence for multi-modal objective, and hence coefficients are varied during the run to have a good balance of exploration and exploitation [8, 12, 13]. A linearly decreasing inertia weight (LDIW-PSO) is used to have greater weight of current particle position during initial iterations, and vice versa.
\begin{equation}
\omega = \omega_{max} - \frac{t}{t_{max}} (\omega_{max}-\omega_{min}) 
\end{equation}
where $\omega$ is inertia weight, $\omega_{max}$ and $\omega_{min}$ are maximum and minimum values respectively for inertia, $t$ is current iteration number and $t_{max}$ is maximum iteration. Time-varying acceleration coefficients (TVAC-PSO) are used to have greater weight of personal best and lesser weight of global best during the initial iterations, and vice versa [9, 16, 21, 22]. In addition to decreasing inertia weight, two acceleration coefficients are also changed in TVAC-PSO as:
\begin{equation}
c_1 = c_1^{max} - \frac{t}{t_{max}} (c_1^{max}-c_1^{min}) 
\end{equation}
\begin{equation}
c_2 = c_2^{min} + \frac{t}{t_{max}} (c_2^{max}-c_2^{min}) 
\end{equation}
where $c_1^{max}$ and $c_1^{min}$ are maximum and minimum values respectively of cognitive acceleration coefficient, and $c_2^{max}$ and $c_2^{min}$ are maximum and minimum values respectively of social acceleration coefficient. Canonical PSO based learning suffers from "oscillation" when personal best and global best are in two opposite directions of current position. As particle moves towards one, distance from other increases and so it moves along opposite direction in next iteration. By keeping higher values of $c_1$ in the early exploration phase and $c_2$ in late exploitation phase, this oscillatory behaviour is minimized.

Multiple swarms are used to ensure that algorithm does not converge prematurely around a local optima. A two-stage PSO (2SPSO) is proposed with multiple independent swarms in first stage, which then collapse into a single swarm in second stage [17, 24]. Particle velocity equation in the first stage of 2SPSO is given as:
\begin{equation}
v_{i,j}^d = \omega v_{i,j}^d + c_1r_1^d(pbest_{i,j}^d-x_{i,j}^d) + c_2r_2^d(sbest_j^d-x_{i,j}^d) \\
\end{equation}

where $v_{i,j}^d$ is velocity of particle $i$ from swarm $j$ in dimension $d$, and $sbest_j^d$ is best position in swarm $j$. Multiple modes are effectively explored in the first stage guided by best particle of each swarm, and then convergence is achieved in the second stage guided by global best. This approach is extended in multi-stage, multi-swarm PSO (MS2PSO) where multiple swarm of the first stage recursively collapse into a single swarm in the final stage [10, 14, 15, 18]. In this study, 8 swarms used in the first 25 iterations collapse into 4 swarms for the next 25 iterations, which further collapse into 2 swarms for the next 25 iteration. After 75 iteration of the first three stages, two remaining swarms collapse into one for last 50 iterations. This approach ensures that multiple modes are explored in the beginning while diversity is retained longer, and thus premature convergence is mitigated.

\section{Results}
Different versions of PSO algorithm discussed above are used to jointly optimize well placement and control settings for the Olympus benchmark. This optimization problem has total 90 control variables which are real valued and vary within a specified range. Population size of 40 was used for total 125 iterations, and thus requiring 5000 simulations for one optimization run. Every algorithm is run twice to discount the impact of random initialization, and better understand impact of micro level rules on meso level interactions. For a high-dimensional and multi-modal problem, random initialization has relatively high impact on search as PSO does not have any mutation operator to aid in exploration.
\begin{figure}[b]
  \centering
  \includegraphics[width=\textwidth]{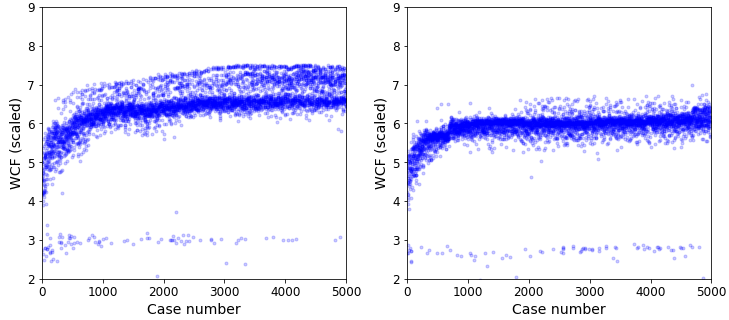}
  \caption{Optimization runs with canonical PSO}
  \label{pso_1_01}
\end{figure}
\begin{figure}[t]
  \centering
  \includegraphics[width=\textwidth]{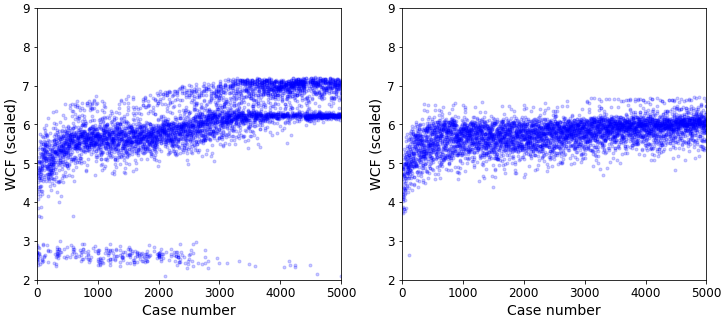}
  \caption{Optimization runs with LDIW-PSO}
  \label{pso_1_02}
\end{figure}

\begin{figure}
  \centering
  \includegraphics[width=\textwidth]{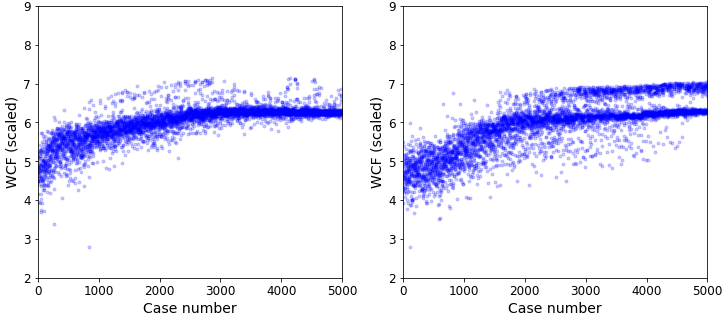}
  \caption{Optimization runs with TVAC-PSO}
  \label{pso_1_03}
\end{figure}

Scaled weighted cumulative fluid (WCF) production after 20 years of simulated production is shown in Figure~\ref{pso_1_01} for two optimization runs with canonical PSO. Scaled WCF value for default case is 4.97, thus it can be concluded that even canonical PSO algorithm works quite well for such high dimensional problem. Figures~\ref{pso_1_02} and ~\ref{pso_1_03} show results from two optimization runs each for LDIW-PSO and TVAC-PSO. Decreasing inertia weight in LDIW-PSO enables exploration in early stages while converging population towards global best in the later stages. TVAC-PSO on the other hand has high $c_1$ and low $c_2$ values in the early stages, which restricts social learning and aids in exploration. Social learning component increases as run progresses and population moves towards global best. 

\begin{figure}[t]
  \centering
  \includegraphics[width=\textwidth]{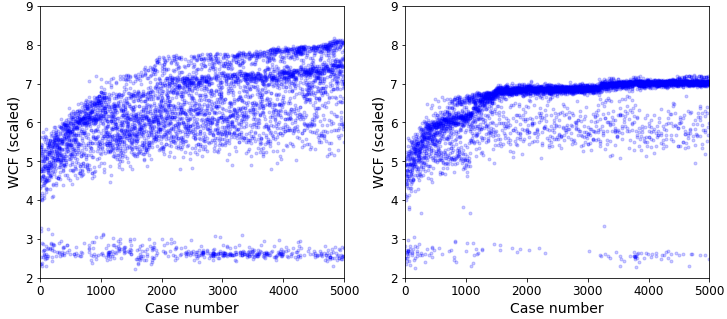}
  \caption{2SPSO optimization runs with $N_s^1$=5 and $N_{t}^1$=25}
  \label{pso_2_01}
\end{figure}

\begin{figure}[t]
  \centering
  \includegraphics[width=\textwidth]{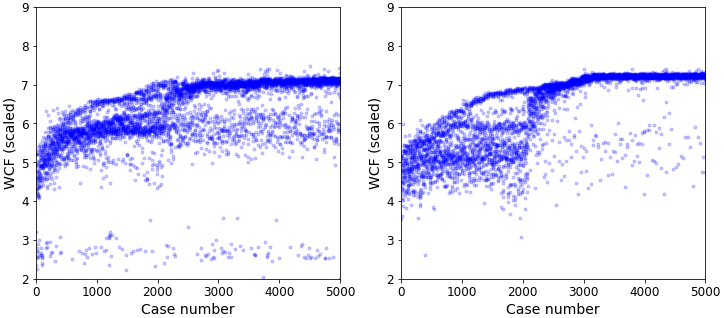}
  \caption{2SPSO optimization runs with $N_s^1$=5 and $N_{t}^1$=50}
  \label{pso_2_02}
\end{figure}

\begin{figure}[t]
  \centering
  \includegraphics[width=\textwidth]{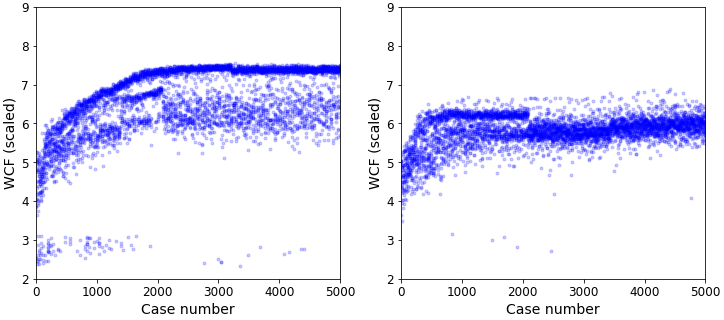}
  \caption{2SPSO optimization runs with $N_s^1$=2 and $N_{t}^1$=50}
  \label{pso_2_03}
\end{figure}

Next, two-stage PSO (2SPSO) is used in which population is divided into a number of independent swarms in the first stage. This aids in exploring multi-modal objective space and prevents premature convergence to a local optima. The swarms collapse into one group in the second stage, and run finally converges towards global best. Inertia weight and acceleration coefficients are kept constant, while number of swarms ($N_s^1$) and number of iterations ($N_{t}^1$) in the first stage are varied to understand their impact on the run behaviour (Figures~\ref{pso_2_01}, ~\ref{pso_2_02} and ~\ref{pso_2_03}). As $N_{t}^1$ increases from 25 to 50 while keeping $N_s^1$ constant, swarms are better able to explore their local structure before converging to global best in the second stage. On the other hand, shorter first stage enables population to better explore around the global best. As $N_s^1$ decreases from 5 to 2 while keeping $N_{t}^1$ constant, the number of samples in each swarm increase which assists in better local search during the first stage while the diversity in the second stage is reduced.

\begin{figure}[t]
  \centering
  \includegraphics[width=\textwidth]{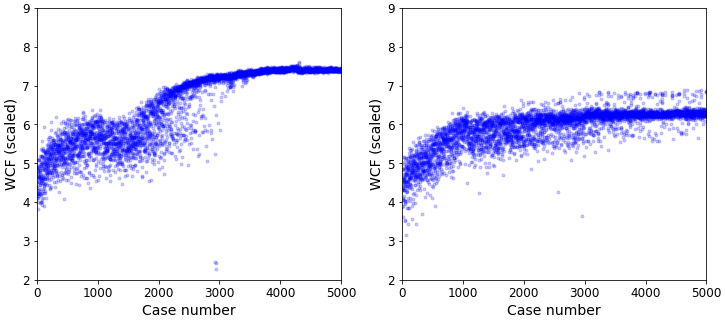}
  \caption{TVAC-2SPSO optimization runs with $N_s^1$=5 and $N_{t}^1$=25}
  \label{pso_3_01}
\end{figure}

\begin{figure}[t]
  \centering
  \includegraphics[width=\textwidth]{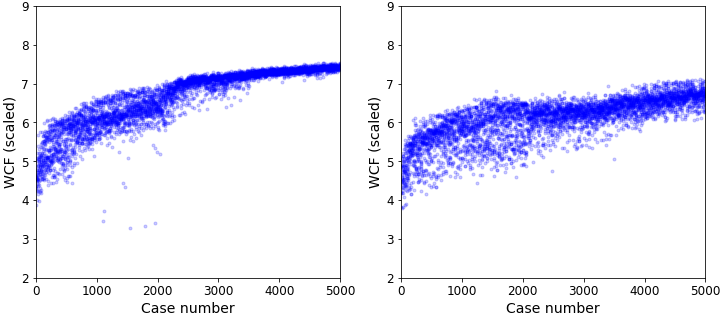}
  \caption{TVAC-2SPSO optimization runs with $N_s^1$=5 and $N_{t}^1$=50}
  \label{pso_3_02}
\end{figure}

Next, two-stage PSO is run with time-varying acceleration coefficients (TVAC-2SPSO) to get the combined effect (Figures~\ref{pso_3_01} and ~\ref{pso_3_02}). Multiple swarms enable better exploration of local structure in first stage, while time-varying coefficients ensures that diversity is maintained even towards the later stage of run.

\begin{figure}[t]
  \centering
  \includegraphics[width=\textwidth]{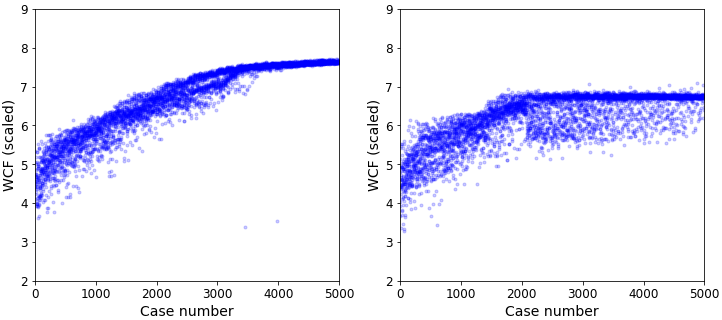}
  \caption{MS2PSO optimization runs with 8 initial swarms, 4 stages and time-varying acceleration coefficients}
  \label{pso_3_03}
\end{figure}

Finally, multi-stage multi-swarm PSO (MS2PSO) is used where multiple swarms of first stage recursively collapse into one swarm in the final stage. Optimization runs are made with 8 initial swarms, 4 stages, and time-varying acceleration coefficients (Figure~\ref{pso_3_03}). As expected, diversity is maintained throughout the run which enables continuous learning and mitigates premature convergence. Social interaction at meso scale is clearly influenced by rules at micro level, and is observable in the progression of optimization run. It may be concluded that even though search rules for a single particle are quite simple, search behaviour of population is quite complex which in turn leads to swarm intelligence.

\section{Conclusions}
High dimensional and multi-modal optimization problem of joint well placement and control is solved using PSO algorithm. Rules at micro level lead to social interactions at meso level, which further lead to collective behaviour at macro level. Coefficients of canonical PSO are tuned in LDIW-PSO and TVAC-PSO, which helps in mitigating issues like premature convergence and oscillation. Multi-stage, multi-swarm approach is used to further improve exploration by maintaining diversity between independent swarms. Two-stage PSO (2SPSO) where multiple swarms of first stage collapse into one in second stage, and multi-stage multi-swarm PSO (MS2PSO) where swarms recursively collapse into one are proposed. These algorithms modify the social behaviour at meso scale based on number of swarms, number of stages and iterations in each stage.

\begin{ack}
Research was made possible by generous cloud credits provided by Amazon Web Services, DigitalOcean and Google Cloud. Olympus benchmark reservoir model was made available by the Netherlands Organization for Applied Scientific Research (TNO).
\end{ack}

\section*{References}

\medskip

\small

[1] Udy, J., Hansen, B., Maddux, S., Petersen, D., Heilner, S., Stevens, K., Lignell, D.\ \& Hedengren, J.D.\ (2017) Review of field development optimization of waterflooding, EOR, and well placement focusing on history matching and optimization algorithms. {\it Processes} {\bf 5}(34).

[2] Fonseca, R.M., Rossa, E.D., Emerick, A.A., Hanea, R.G.\ \& Jansen, J.D.\ (2020) Introduction to the special issue: overview of OLYMPUS optimization benchmark challenge. {\it Computational Geosciences}, {\bf 24} (6):1933-1941. 

[3] Tanaka, S., Wang, Z., Dehghani, K., He, J., Velusamy, B.\ \& Wen, X.\ (2018) Large scale field development optimization using high performance parallel simulation and cloud computing technology. {\it SPE Annual Technical Conference and Exhibition}.

[4] Wang, D., Tan, D.\ \& Liu, L.\ (2018) Particle swarm optimization algorithm: an overview. {\it Soft Computing}, {\bf 22}:387-408.

[5] Sengupta, S., Basak, S.\ \& Peters, R.A.\ (2019) Particle swarm optimization: a survey of historical and recent developments with hybridization perspectives. {\it Machine Learning and Knowledge Extraction}, {\bf 1}:157-191.

[6] Zhang, Y., Wang, S.\ \& Ji, G.\ (2015) A comprehensive survey on particle swarm optimization algorithm and its applications. {\it Mathematical Problems in Engineering}, {\bf 2015}.

[7] Baumann, E.J.M, Dale, S.I.\ \& Bellout, M.C.\ (2020) FieldOpt: A powerful and effective programming framework tailored for field development optimization. {\it Computers and Geosciences}, {\bf 135}.

[8] Clerc, M.\ \& Kennedy, J.\ (2002) The particle swarm - explosion, stability, and convergence in a multidimensional complex space. {\it IEEE Transactions on Evolutionary Computation}, {\bf 6} (1):58-73.

[9] Zhan, Z-H., Zhang, J., Li, Y.\ \& Chung, H.S-H.\ (2009) Adaptive particle swarm optimization. {\it IEEE Transactions on Systems Man, and Cybernetics — Part B: Cybernetics}, {\bf 39} (6):1362-1381.

[10] Du, W., Gao, Y., Liu, C., Zheng, Z.\ \& Wang, Z.\ (2015) Adequate is better: particle swarm optimization with limited-information. {\it Applied Mathematics and Computation}, {\bf 268}:832-838.

[11] Ding, S., Jiang, H., Li, J.\ \& Tang, G.\ (2014) Optimization of well placement by combination of a modified particle swarm optimization algorithm and quality map method. {\it Computational Geosciences}, {\bf 18}:747–762.

[12] Arasomwan, M.A.\ \& Adewumi, A.O.\ (2013) On the Performance of Linear Decreasing Inertia Weight Particle Swarm Optimization for Global Optimization. {\it The Scientific World Journal}, {\bf 2013}.

[13] Harrison, K.R., Engelbrecht, A.P.\ \& Ombuki-Berman, B.M.\ (2016) Inertia weight control strategies for particle swarm optimization. {\it Swarm Intelligence}, {\bf 10}:267–305.

[14] Piotrowski, A.P., Napiorkowski, J.J.\ \& Piotrowska, A.E.\ (2020) Population size in Particle Swarm Optimization. {\it Swarm and Evolutionary Computation}, {\bf 58}.

[15] Liang, J.J., Qin, A.K., Suganthan, P.N.\ \& Baskar, S.\ (2006) Comprehensive learning particle swarm optimizer for global optimization of multimodal functions. {\it IEEE Transactions on Evolutionary Computation}, {\bf 10} (3):281-295.

[16] Isiet, M.D.\ (2019) {\it Self-adapting control parameters in particle swarm optimization}. MS Thesis, The University of British Columbia.

[17] Zhao, Q.\ \& Li, C.\ (2020) Two-stage multi-swarm particle swarm optimizer for unconstrained and constrained global optimization. {\it IEEE Access}, {\bf 8}:124905-124927.

[18] Xia, X., Gui, L.\ \& Zhan, Z.\ (2018) A multi-swarm particle swarm optimization algorithm based on dynamical topology and purposeful detecting. {\it Applied Soft Computing}, {\bf 67}:126-140.

[19] Kachitvichyanukul, V.\ (2012) Comparison of Three Evolutionary Algorithms: GA, PSO, and DE. {\it Industrial Engineering and Management Systems}, {\bf 11} (3):215–223.

[20] Oliveira, M., Pinheiro, D., Macedo, M., Bastos-Filho, C.\ \& Menezes, R.\ (2020) Uncovering the social interaction network in swarm intelligence algorithms. {\it Applied Network Science}, {\bf 5} (24).

[21] Chih, M., Lin, C., Chern, M.\ \& Ou, T.\ (2014) Particle swarm optimization with time-varying acceleration coefficients for the multidimensional knapsack problem. {\it Mathematical Modelling}, {\bf 38} (4):1338–1350.

[22] Mohammadi-Ivatloo, B., Rabiee, A., Soroudi, A.\ \& Ehsan, M.\ (2012) Iteration PSO with time varying acceleration coefficients for solving non-convex economic dispatch problems. {\it International Journal of Electrical Power \& Energy Systems}, {\bf 42} (1):508-516.

[23] Mamei, M., Menezes, R., Tolksdorf, R.\ \& Zambonelli, F. (2006) Case studies for self-organization in computer science. {\it Journal of Systems Architecture}, {\bf 52} (8-9):443-460.

[24] Xu, G., Cui, Q., Shi, X., Ge, H., Zhan, Z., Lee, H.P., Liang, Y., Tai, R.\ \& Wu, C.\ (2019) Particle swarm optimization based on dimensional learning strategy. {\it Swarm and Evolutionary Computation}, {\bf 45}:33-51.

\end{document}